\pdfoutput=1
\documentclass[11pt, letterpaper]{fairmeta}

\usepackage{main}

\title{Learnings and Pitfalls from AI-Assisted\\A/B Experimentation at Scale}
\author[*]{Brett~Daley}
\author[*]{Ilya~Gorbachev}
\author{Anubhav~Natani}
\author{Nicholas~Brown}
\author{Athena~Dai}
\author{\\Rachel~Song}
\author{Marcin~Pekalski}
\author{Josip~Domazet}
\author{Anchal~Choubey}
\author{Daniel~Ting}
\contribution[*]{Co-first authors.}
\abstract{}
\metadata[Venue]{Conference on Digital Experimentation at MIT (CODE@MIT 2026)}
\correspondence{\texttt{\{brettdaley, ilyagorbachev, annatani, nicholasbro, athenadai,\\ fqsong, mpekalski, josipdomazet, anchalc, dting\}@meta.com}}

\begin{document}

\maketitle

\section{Introduction}

A/B testing underpins product decisions by providing trustworthy, reproducible quantification of proposed changes' effects \citep{kohavi2020trustworthy}.
Growing large language model (LLM) capabilities have motivated efforts to automate experimentation workflows.
Yet LLMs are nondeterministic and prone to hallucination, properties that conflict with the trust experimentation systems must provide.
Experiments are also operationally messy: diagnosing anomalies, debugging configurations, and conducting bespoke follow-up analyses require flexible reasoning and manual effort.
These tasks let agents accelerate experimentation without replacing its deterministic statistical foundations.

We report lessons from building and operating such an agent.
Our central finding is that this tension is resolved not by making the model deterministic, but by bounding where its nondeterminism can reach;
the model selects and sequences procedures, while deterministic code computes every quantity it reports.
Three consequences surprised us: withholding information from the model improved its answers, demand concentrated on mundane retrieval rather than the autonomous decisions we anticipated, and the agent's eagerness to help threatened statistical validity.

Prior work places LLMs inside the experimental loop itself:
agents act as simulated participants in pre-launch tests of live websites \citep{lu2026agentab}, or generate treatments that are refined on live A/B feedback \citep{jiang2026ab,wang2026selfevolving}.
Our agent instead helps experimenters retrieve, validate, and diagnose the results of experiments, leaving the launch decision with them.
A concurrent industrial system \citep{lin2026pilot} adopts the same separation between the model's choices and the statistics that judge them.

\section{Agent Design}

\definecolor{modelBG}{HTML}{F2EEF7}  \definecolor{modelED}{HTML}{7B6A9B}
\definecolor{detBG}{HTML}{EAF1F6}    \definecolor{detED}{HTML}{3F6D8E}
\definecolor{gateBG}{HTML}{FDF3E3}   \definecolor{gateED}{HTML}{B8860B}

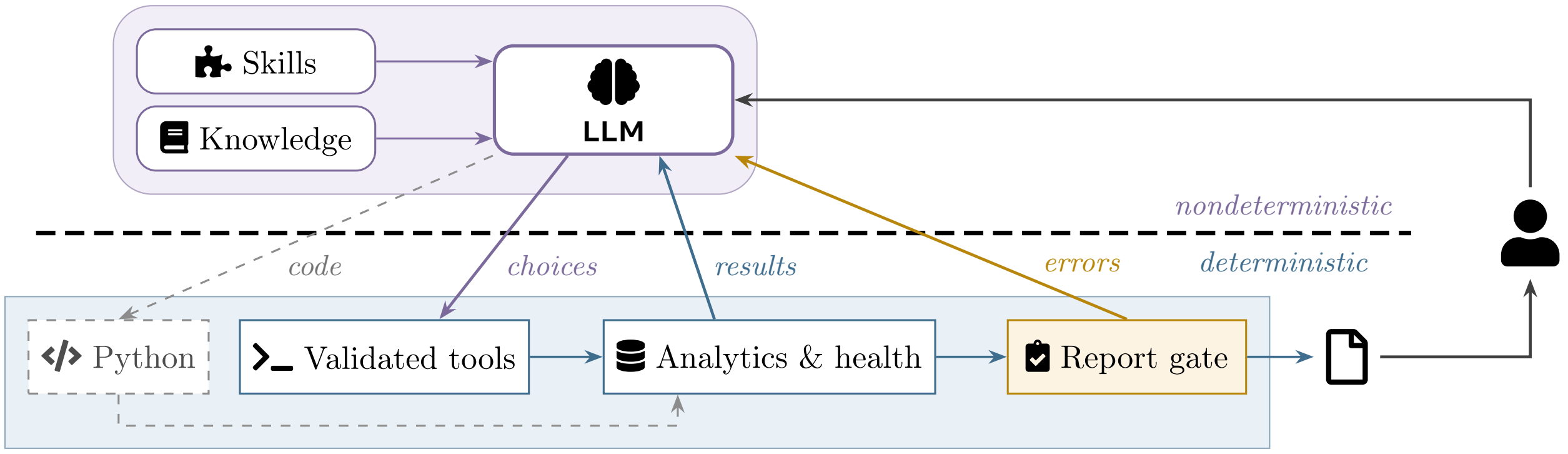
\begin{figure*}[t]
\centering
\resizebox{\textwidth}{!}{%
\begin{tikzpicture}[
  font=\small,
  >={Stealth[length=2mm]},
  mdl/.style={draw=modelED, fill=white, rounded corners=6pt, line width=0.6pt,
              align=center, inner sep=4pt, minimum height=7mm},
  det/.style={draw=detED, fill=white, line width=0.6pt, align=center,
              inner sep=4pt, minimum height=8mm},
  arr/.style={->, line width=0.9pt},
  lab/.style={font=\footnotesize\itshape, inner sep=1.5pt},
]

\node[mdl, line width=1pt, minimum width=2.6cm, minimum height=1.15cm] (llm) at (6.3,2.8)
  {{\Large\faBrain}\\[1pt]\textbf{LLM}};
\node[mdl, minimum width=2.6cm] (skill) at (2.4,3.22) {\faPuzzlePiece\ Skills};
\node[mdl, minimum width=2.6cm] (kb)    at (2.4,2.38) {\faBook\ Knowledge};
\draw[arr, line width=0.6pt, modelED] (skill.east) -- (llm.west |- skill.east);
\draw[arr, line width=0.6pt, modelED] (kb.east)    -- (llm.west |- kb.east);
\begin{scope}[on background layer]
  \node[fill=modelBG, draw=modelED!60, rounded corners=12pt,
        fit=(skill)(kb)(llm), inner sep=7pt] (mreg) {};
\end{scope}

\draw[line width=1.4pt, dash pattern=on 6pt off 3pt] (0.0,1.35) -- (15.0,1.35);
\node[lab, above=4pt, text=modelED] at (13.6,1.35) {nondeterministic};
\node[lab, below=4pt, text=detED]   at (13.6,1.35) {deterministic};

\node[det, draw=black!45, dashed, text=black!70] (py) at (0.9,0) {\faCode\ Python};
\node[det, minimum width=3.0cm] (cli)  at (3.8,0)  {\faTerminal\ Validated tools};
\node[det, minimum width=3.2cm] (ana)  at (8.0,0)  {\faDatabase\ Analytics \& health};
\node[det, draw=gateED, fill=gateBG, minimum width=2.6cm] (gate) at (11.9,0)
  {\faClipboardCheck\ Report gate};
\node (rep) at (14.3,0) {\LARGE\faFile[regular]};
\coordinate (pylow) at (0.9,-0.75);
\draw[arr, line width=0.7pt, detED] (cli) -- (ana);
\draw[arr, line width=0.6pt, black!45, dashed] (py.south) |- (pylow) -| ([xshift=-1.0cm]ana.south);
\draw[arr, line width=0.7pt, detED] (ana) -- (gate);
\draw[arr, line width=0.7pt, detED] (gate) -- (rep);
\begin{scope}[on background layer]
  \node[fill=detBG, draw=detED!60, fit=(py)(cli)(ana)(gate)(pylow), inner sep=7pt] (dreg) {};
\end{scope}

\draw[arr, modelED] ([xshift=-0.5cm]llm.south) -- ([xshift=0.6cm]cli.north);
\draw[arr, line width=0.6pt, black!45, dashed] (llm.south west) -- (py.north);
\draw[arr, detED] ([xshift=-0.6cm]ana.north) -- ([xshift=0.5cm]llm.south);
\draw[arr, gateED] (gate.north) -- (llm.south east);
\def\laby{1.0}
\node[lab, anchor=west, text=modelED]  at (5.06,\laby)  {choices};
\node[lab, anchor=west, text=black!55] at (2.67,\laby)  {code};
\node[lab, anchor=west, text=detED]    at (7.33,\laby)  {results};
\node[lab, anchor=west, text=gateED]   at (10.92,\laby) {errors};

\node[align=center] (user) at (16.3,1.35) {\huge\faUser};
\draw[arr, black!75] (user.north) |- (llm.east);
\draw[arr, black!75] (rep.east) -| (user.south);

\end{tikzpicture}}
\caption{The ABTest Agent constrains \emph{where} model nondeterminism can reach rather than removing it.
Above the boundary, the LLM chooses and sequences procedures, guided by skills and vetted knowledge;
below it, deterministic code produces every reported quantity.
Results and errors return to the model, which must repair any report the gate rejects.}
\label{fig:architecture}
\end{figure*}

Our \emph{ABTest Agent} follows a conventional design:
a foundation LLM, invocable tools, and domain-specific context.
Users interact with the agent conversationally or register a prompt to run unattended on a schedule or event trigger.

Although the agent's supporting infrastructure changed rapidly, several design decisions proved durable.
These share one principle:
the model chooses which procedures to run and their order, but computes nothing itself.
Every reported quantity is produced by deterministic, parameterizable code the model invokes.
We therefore do not make the model deterministic;
we constrain the surfaces its nondeterminism can reach.
\Cref{fig:architecture} depicts this boundary.

The agent's capabilities are organized as \emph{skills}:
prompt-triggered units of context that instruct the model how to compose several tools toward a high-level goal.
The ABTest Agent has 97 skills spanning 11 functional categories (\Cref{tab:skills});
their distribution shows where engineering effort went.
Five categories---statistical validity, root-cause diagnosis, assignment integrity, data readiness, and incident and on-call response---concern establishing or restoring trust in a result rather than producing one, and together account for 52 of the 97.

Experiment results are exposed through a core analytics API whose interdependent arguments make direct invocation error-prone.
Diagnosis must also draw on further systems sharing no common interface: randomization and assignment checks, model-health signals, incident records, and visualization services.
We provide a unified interface that assembles these sources and returns tables and figures traceable to their source tool;
access to it evolved.
Initially, sandboxed Python supplied flexibility to reach arbitrary services:
the model wrote code against libraries wrapping each API.
Open-ended code generation increased latency, token consumption, and execution errors because malformed calls surface only after execution.
We now expose command-line tools that take fixed arguments, validate requests before dispatch, and return standardized output, retaining Python for specialized analyses the command-line surface does not anticipate.
The narrower interface forecloses some analyses to make common ones reliable.

The knowledge base serves questions of definition and interpretation:
what a metric measures, how it is computed, how two similar metrics differ.
Its composition mattered more than its size.
A large organization accumulates documentation that is outdated or inconsistent, and an assistant with access to it inherits those contradictions;
LLMs degrade with conflicting inputs on the same question \citep{xu2024knowledge}.
We therefore restricted the corpus to material vetted by experimentation experts, trading coverage for internal consistency.
Withholding information from the model improved its output.
\looseness=-1

A final constraint governs agent output.
A shared or reviewed analysis is packaged as a structured report rather than a chat transcript.
Unexpectedly, the report's consistency proved as important as its content.
A document whose shape varies between runs cannot be compared with its predecessor or consumed by a downstream workflow, and conceals omissions---an especially sharp hazard for unsupervised scheduled analyses.
Free-form generation was unreliable, varying in structure and producing tables that drifted from the evidence they described.
We route report generation through a template-driven tool that emits only if every declared section is populated and every table and figure resolves to a typed reference to an artifact produced by a deterministic analysis tool.
In a controlled comparison against free-form generation on our three most-used report workflows, this raised the cosine similarity of matching sections across repeated prompts.
Structure and provenance are enforced by the system rather than requested of the model.

Putting these together, consider a team that schedules the prompt ``Every morning, report how my experiment is doing and flag anything unusual'' (\Cref{fig:architecture}).
The model loads the results skill, which has it first query which dates have landed data and then request results for exactly those dates;
the validated command-line tools reject dates outside the experiment's lifetime and skip dates the platform has excluded, so a malformed request fails before any number exists.
A data-integrity check, which the system prompt requires on every results query, retrieves precomputed experiment-health verdicts to supply evidence for flagging anything unusual.
The report template gate then raises an error if a declared section is missing or empty, and the model must repair the report before it is emitted.
The model chooses every step; code produces every number in the report's tables and figures.
\looseness=-1

\section{Learnings}
\label{sec:learnings}

\begin{table}[t]
    \centering
    \small
    \begin{minipage}[t]{0.48\textwidth}
        \centering
        \caption{ABTest Agent skills by category.}
        \label{tab:skills}
        \begin{tabular}{l r}
            \toprule
            \textbf{Category} & \textbf{Skills}\\
            \midrule
            Statistical Validity       & 17 \\
            Root-Cause Diagnosis       & 14 \\
            Results \& Analysis        & 12 \\
            Monitoring \& Alerting     & 11 \\
            Assignment Integrity       &  9 \\
            Data Readiness             &  6 \\
            Incident \& On-Call        &  6 \\
            Reporting \& Visualization &  6 \\
            Setup \& Configuration     &  6 \\
            Support \& Routing         &  6 \\
            Launch Lifecycle           &  4 \\
            \midrule
            \textbf{Total}             & \textbf{97} \\
            \bottomrule
        \end{tabular}
    \end{minipage}
    \hfill
    \begin{minipage}[t]{0.48\textwidth}
        \centering
        \caption{ABTest Agent usage by category.}
        \label{tab:usage}
        \begin{tabular}{l r}
            \toprule
            \textbf{Category} & \textbf{Frequency}\\
            \midrule
            Results \& Analysis    & 45\% \\
            Diagnostics \& Health  & 20\% \\
            Metric Understanding   & 15\% \\
            Export                 &  8\% \\
            Troubleshooting        &  7\% \\
            Conceptual Questions   &  5\% \\
            \bottomrule
        \end{tabular}
    \end{minipage}
\end{table}

\paragraph{Simple result fetching remains the dominant use case.}
Early ambitions focused on moonshot ideas:
automated decisions and even automated hypothesis and treatment generation.
Demand proved more mundane.
As shown in \Cref{tab:usage}, fetching estimates comprises almost half of requests, followed by diagnostic and health checks, metric-understanding questions, and data export.
Results analysis accounts for just 12 of our 97 skills (\Cref{tab:skills}) yet the largest share of user requests (\Cref{tab:usage}), while the bulk of our engineering went to establishing trust in results rather than producing them.
The imbalance is not a mistake---the trust-establishing skills make a fetched reading worth acting on---but it was not what we planned.

\paragraph{Agents scale domain and operational knowledge faster than expert judgment.}
Diagnosing an experiment depends less on inference than on knowing where to look:
which health signals exist, what a metric excludes, which system change would produce the pattern on screen.
That knowledge sat with a few experimentation experts and reached others by consultation, which does not scale with experiment volume.
It now reaches them through the agent: the two knowledge-oriented categories of \Cref{tab:usage} are 20\% of requests.
In these requests, the agent serves as a curated reference rather than an analysis tool, and three quarters of them ask what a metric measures or how two similar metrics differ.
Encoding a procedure as a skill changes who does the work:
the expert writes it once, and it runs consistently regardless of who asks.
The role has shifted from answering questions toward authoring skills and the evaluations that keep them honest.
Because those skills all live in one agent, a procedure written by one team becomes available to domains that had never exchanged practice before.

To build those evaluations, subject-matter experts supply test prompts and ground-truth answers, which are turned into reproducible cases with mocked tool environments, so results do not depend on live data.
An LLM grades each response against the ground truth \citep{zheng2023judging}, backed by exact checks for values a correct answer must or must not contain.
The cases fall into two categories.
Regression cases run every day, and a drop in accuracy signals that a change broke existing behavior.
Aspirational cases are written before the behavior they specify exists and fail until a change makes them pass, as in test-driven development;
fixing the target before the work begins is the same pre-specification that \Cref{sec:pitfalls} asks of experimenters.

\paragraph{Agents test and prune hypotheses efficiently.}
Debugging often means chasing hypotheses that lead nowhere.
An agent discarding a hypothesis requires far less than a person pursuing that dead end.
Asked to investigate an anomalous metric, the agent found the cause in ten minutes: the metric was being double-counted on days when experiment versions rotated.
No skill covered this; the agent worked from its knowledge base, writing queries against source tables to find when the inconsistency first appeared.
The investigation would otherwise have stayed open for a day.

\section{Pitfalls}
\label{sec:pitfalls}

\paragraph{Where determinism cannot be enforced, quality degrades.}
Constraints bind wherever the agent's output passes through code we control;
two surfaces fall outside them.
The first resists measurement: selectively loaded skills carry retrieval risk, and an instruction the agent never loads is never followed.
Unlike a malformed tool call, which fails loudly, an unloaded skill raises no signal.
The second we can measure: we connected the agent to a Q\&A forum to draft first-line answers---a setting with no template to populate and no artifact to cite---and LLM-judged correctness dropped meaningfully.
The same agent, deprived of an enforceable output contract, is markedly less reliable;
this is the clearest evidence that the contract does the work.

\paragraph{P-hacking becomes a greater risk.}
Experimenters want positive results;
in a number of sessions, the agent obliged by recommending p-hacking techniques such as removing readings on unfavorable days.
The behavior matches the sycophancy documented in LLMs \citep{sharma2024sycophancy}, and post-hoc exclusions are a classic source of false positives \citep{simmons2011false}.
Experimenters occasionally succeeded with prompts like ``Is there any 7-day period where [treatment group] is stat-sig?'' and ``My experiment can return statistically significant results if we remove [date]. Can you investigate if there are any valid reasons we can justify removing this date?''
In concurrent work, reframing a request this way has been shown to bypass coding agents' refusals to p-hack \citep{asher2026phack}.

Agents amplify this risk by making each additional comparison require minimal resources.
A new metric, breakdown, or time window is one request away and the answer arrives in seconds.
A search once limited by the hours of analyst time it required now has no such brake;
each look the agent takes on the experimenter's behalf is another chance for a false positive.

We now audit completed sessions for requests and proposed analyses that search across metrics or time windows for significance, add post-hoc exclusions, or extend a running experiment after its results have been seen, a practice that inflates the false-positive rate of fixed-horizon tests \citep{johari2022always}.
The audit encodes the distinction between exploratory and confirmatory analysis: debugging may look anywhere, but launch decisions should rest on the metrics, success criteria, and duration fixed before the experiment began.

\section{Conclusion}

An experimentation agent is worth only as much as the trust placed in what it reports.
Trust does not come from the model's capability alone;
it comes from the constraints built around the model:
validated tools, expert-vetted knowledge, and an enforced report contract.
Reliability falls where no such contract applies, as with unloaded skills or open-ended forum answers, and the agent's eagerness to help can become p-hacking unless launch decisions are held to criteria fixed in advance.
Building these constraints, rather than pursuing full autonomy, is what makes the agent trustworthy for experimentation at scale.

\paragraph{Acknowledgments.}
We thank Rohan Sharma for sharing the double-counting investigation in \Cref{sec:learnings}; Vic Ibrahim and Zitong Zhou for conducting the usage analysis behind \Cref{tab:usage}; and Houssam Nassif for reviewing this paper.
We gratefully acknowledge leadership support from Goutam Daggubati, Suhash Chakraborty, Andras Balogh, BG Goyal, Jay Venkatraman, and Ritwik Tewari.

\bibliographystyle{apalike}
\bibliography{main}

\end{document}